\documentclass[preprint,authoryear,12pt]{elsarticle}

\usepackage{amssymb}

\journal{Neuroimage}
\usepackage{amssymb,amsmath}
\usepackage[normalem]{ulem}
\usepackage[color,final]{showkeys}
\usepackage{multirow}
\usepackage{xhfill}
\usepackage{graphics}
\usepackage{graphicx}
\usepackage{subfigure}
\usepackage{epsfig}
\usepackage{url}

\newcommand{\comm}[1]{{\color{red}(#1)}}
\newcommand{\revise}[1]{{\color{blue}#1}}

\def\ni{\noindent}

\begin{document}

\begin{frontmatter}



\title{Randomized Structural Sparsity via Constrained Block Subsampling for  Improved Sensitivity of Discriminative Voxel Identification}
\author[label1,label2,label4]{Yilun Wang}
\author[label2]{Junjie Zheng}
\author[label1]{Sheng Zhang}
 \address[label1]{School of Mathematical Sciences, University of Electronic Science and Technology of China, Chengdu,
Sichuan, 611731 P. R. China.}
\author[label2]{Xunjuan Duan}
\author[label2]{Huafu Chen \corref{cor1}}
  \address[label2]{Key laboratory for Neuroinformation of Ministry of Education,
School of Life Science and Technology, and Center for Information in Biomedicine,
University of Electronic Science and Technology of China,
Chengdu, Sichuan, 611054, P. R. China.}
\address[label4]{Center for Applied Mathematics,
Cornell University, Ithaca, NY, 14853, USA}
 \ead{chenhf@uestc.edu.cn}
 \cortext[cor1]{Corresponding Author}


\begin{abstract}
In this paper, we consider voxel selection for functional Magnetic Resonance Imaging (fMRI) brain data with the aim of finding a more complete set of  probably correlated discriminative voxels, thus improving interpretation of the discovered potential biomarkers. The main difficulty in doing this is an extremely high dimensional voxel space and  few training samples, resulting in unreliable feature selection. 
In order to deal with the difficulty, stability selection has  received a great deal of attention lately, especially due to its
finite sample control  of false discoveries and
transparent principle for choosing  a proper amount of regularization.
However, it fails to make explicit use of the correlation property or structural information 
of these discriminative features and leads to large false negative rates. In other words,   many relevant but probably correlated discriminative voxels are missed. Thus,  we propose a new variant on stability selection  ``randomized structural sparsity", which incorporates the idea of structural sparsity. 
%
%
Numerical experiments
 demonstrate that our method can be superior in controlling for  false negatives while also keeping the control of false positives inherited from stability selection. 
\end{abstract}


\end{frontmatter}


\section{Introduction} \label{Sec:intro}
\subsection{Problem Statement}
 Decoding neuroimaging data, also called brain reading, is a kind of pattern recognition  that has led to impressive results, such as  guessing which image a subject is looking at
from his brain activity \citep{Haxby01MVA}, as well as in medical diagnosis, e.g., finding out whether a person is a healthy control or a patient. 
%
This pattern recognition  typically consists of two important components: feature selection and classifier design.
While the predictive or classification accuracy of these designed classifiers have received most attention in most existing literature, 
feature selection 
is an even  more important goal in many practical applications including  medical diagnosis  where selected voxels  can be used as biomarker candidates \citep{Guyon03Introduction}.

However,
most traditional feature selection methods fail to  discover in a stable manner the ``complete" discriminative features accurately. They mainly aim to construct a concise classifier and they often  select only a
minimum subset of features, ignoring those correlated or redundant but informative features \citep{Guyon03Introduction,Blum1997245}. In addition,  the stability of the selected features is often ignored \citep{buhlmann2011statistics,Cover65Pattern}, because the inclusion of some noisy features or the exclusion of some informative features may not affect the prediction accuracy \citep{Yu08denseFeatureGroups}, which is their main objective. Therefore, a large number of uninformative, noisy voxels that do
not carry useful information about the category label, could be included in the final feature detection results \citep{Langs2011497}, while some informative, possibly redundant features might be missed.

In this paper, we focus on feature selection on  functional MRI (fMRI) data where each voxel  is considered as a feature. These features are often correlated or redundant. We focus on the ``completeness"  and ``stability" of feature selection, i.e. aim to discover as many as possible  informative but possibly redundant features  accurately and stably, in contrast to most of the existing methods which mainly aim to find a subset of discriminative features which are expected to be uncorrelated. This way,  potential biomarkers revealed by the discovered discriminative voxels, in both cognitive tasks and  medical diagnoses are expected to be more credible. 

\subsection{Advantages and Limitations of Sparse priors in Multivariate  Neuroimaging Modeling}

There are in general three main categories of supervised feature selection algorithms:  filters, embedded methods, and wrappers \citep{Guyon03Introduction}. The filter methods usually separate feature section from classifier development. For example,  Fisher Score \citep{Duda:2000:PC:954544},  is among the most representative algorithms in this category. The wrapper methods use a predictive model to score feature subsets. Each new subset is used to train a model, which is tested on a hold-out set, and the features are scored according to their predictive power.  The embedded models perform feature selection during learning. In other words, they achieve model fitting and feature selection simultaneously. 
The following sparsity related feature selection models are all typical embedded methods, which we will mainly focus on in this paper.  

In this paper,  we consider  commonly used  supervised learning to identify the discriminative brain voxels from given training fMRI data. 
While the  classification problem is considered most often, the regression problem can be treated in a similar way.  We consider the following linear model. 
  \begin{equation} \label{eq:linearmodel}
 \mathbf{y} = \mathbf{X}\mathbf{w} + \epsilon
 \end{equation}
 where $\mathbf{y} \in \mathbb{R}^{n\times 1}$ is the binary classification  information and $\mathbf{X} \in \mathbb{R}^{n\times p}$ is the given training fMRI data and $\mathbf{w} \in \mathbb{R}^{p\times 1}$ is the unknown weights reflecting the
 degree of importance of each voxel. As  a multivariate inverse inference problem, identification of discriminative voxels  is based on the values of the weight vector $\mathbf{w}$ and their importance is  proportional to the absolute values of the components. Therefore, feature selection is also called support identification in this context, because the features corresponding to the nonzero $\mathbf{w}$ components are considered as the relevant features. 


Considering that the common challenge in this field is the curse of dimensionality $p \gg n$,  
we are focusing on  sparsity-based voxel selection methods, because 
sparsity is motivated by the prior knowledge that  the most discriminative  voxels are only
a small portion of the whole brain voxels \citep{Yamashita08SparseEstimation}. 
However, sparsity alone is not
sufficient for making reasonable and stable inferences. 
Plain sparse learning models often provide
overly sparse and hard-to-interpret solutions where 
the  selected voxels are  often scattered
\citep{Rasmussen2012SparsityFMRI}, though they might be useful if a concise classifier is expected. Specifically, if there is a set of  highly correlated features, then only a small portion of representative voxels are
selected, resulting into a large false negative rate and a potential biomarker that is hard to trust.   
In addition, let us denote  the support of the true sparse vector $\bar{\mathbf{w}}$ as $S$, and  the number of its nonzeros  as $\ell$. For the success of finite sample recovery by the plain $\ell_1$ norm regularized model, $\ell$ should be smaller than $n$. Let subsets of the columns of the design matrix $\mathbf{X}$ larger than $\ell$ must be well conditioned. In particular, the design matrix $\mathbf{X}_S$  should be sufficiently well conditioned and should not be too correlated to the columns of $\mathbf{X}$ corresponding to the noisy subspace $\mathbf{X}_{\bar{S}}$ \citep{Varoquaux12Clustering}.

Thus we have to extend the plain sparse learning model to incorporate important structural features of brain imaging data,  such as brain segregation and integration, in order to achieve stable, reliable and interpretable results.

\subsection{Existing  Extensions of the Plain Sparse Model}

As mentioned above,  two common hypotheses have been made for fMRI data analysis: sparsity and compact structure. In sparsity, few relevant and highly  discriminative voxels  are implied in the classification task; in compact structure, relevant discriminative voxels are grouped into several distributed  clusters, and the voxels within a cluster have similar behaviors and are, correspondingly, strongly  correlated. 
Thus making use of these  two hypotheses is very important, and we will review some state-of-the-art existing works in this direction.  

Elastic net regression \citep{Zou05ElsticNet} tries to make use of the voxel correlation by adding an $\ell_2$ regularization, also  called the Tikhonov regularization,  to the classical $\ell_1$ penalty  \citep{RyaliCSM12L1L2Network} to deal with highly correlated features.  
Recently,  other penalties have been added to consider the correlated features besides the Tikhonov regularization \citep{Dubois14TVL2L1}. For example, both $\ell_1$ penalization and Total-Variation (TV) penalization are used simultaneously for voxel selection  \citep{Gramfort:2013:IPR:2552484.2552506}, where the TV penalization is used to make use of the assumption that the activations are spatially correlated and the weights of the voxels are close to piece-wise constant. In addition,  $\ell_2$-fusion penalty can be used if successive regression coefficients are known to vary slowly and can also be interpreted in terms of correlations between successive features in some cases \citep{hebiri2011}.  While these models based on both $\ell_1$ norm and other certain smoothing penalty,  might achieve improved  sensitivity  over the plain $\ell_1$ norm regularized model, they do not  make use of any explicit prior grouping or other structural information of the features \citep{Xia10graph-basedENet}.


Correspondingly, another class of methods  to   make  more explicit use of the segregation and integration of the brain,
is based on  structured sparsity models \citep{Bach12StructuredSparsity,Schmidt11NIPS,chen2012}, which have
been proposed to extend the well-known plain $\ell_1$ norm regularized models by enforcing more structured
constraints on the solution. 
For example, the discriminative voxels are
grouped together into few clusters \citep{Baldassarre12StructuredSparsityFMRI,Michael11TV}, 
where the (possibly overlapping) groups have often  been known as a prior information \citep{Ye12groupNonconvex,Liu10NIPSGroupeTree,Yuan13Overlapping,Jacob09GLO,Ye09L21,Ng11classifier}. 
However, in many cases, the grouping information is not available beforehand, and one can use either the anatomical regions as an approximation \citep{Batmanghelich12Groupsparisty}, or   the data driven methods to obtain the grouping information such as   hierarchical agglomerative clustering (Ward hierarchical clustering, for example) and a top-down step to prune the generated tree of hierarchical clusters in order to obtain the grouping information  \citep{Michel2012SupervisedClustering,jenatton12multiscale}. 

While structural sparsity helps select the correlated discriminative voxels and  is necessary for the ``completeness" of the selected discriminative voxels, the result of feature selection may not be stable and is likely to include many noisy and uninformative voxels.  For years,
the idea of ensemble has been applied  to reduce the variance of feature selection result  \citep{Hastie09Ensemle,DaMota2014203}.  Among them, one important class of methods for high dimensional data analysis is stability selection \citep{Meinshausen10StabilitySelection,Shah13StabilitySelection}. It is an effective way for voxel selection and structure estimation,  based on subsamplings (bootstrapping would behave similarly). 
It aims to alleviate the disadvantage of the plain $\ell_1$ norm regularized model, which either selected by chance non-informative regions, or even worse, neglected relevant
regions that provide duplicate or redundant classification information \citep{Mitchell04LearningDecodeCognitive,Li12SparseLocalization}. This is due in part to the
worrying instability and potential deceptiveness of the most
informative voxel sets when information is non-local or
distributed \citep{Anderson10critiqueMVPA,Poldrack06Criti}. Correspondingly, one major advantage of stability selection is the control of false positives, i.e. it is able to  obtain the selection probability threshold based on the theoretical boundary on the expected number of false positives.  In addition, stability selection is not very sensitive to the choice of the sparsity penalty parameter,  and   stability selection has been applied to the pattern recognition based on brain fMRI data and achieved better results than  plain $\ell_1$ norm regularized models  \citep{Ye12SparseLearningSS,Cao2014Sparse,Ryali20123852,Mairal20131451,Meinshausen20131439,Rondina14}. For example,  SCoRS \citep{Rondina14} is an application of stability selection  designed for the
particular characteristics of neuroimaging data.
Notice that we are focusing on the feature selection here. As for the prediction or classification accuracy, this ensemble or averaging idea has already been applied to reduce the prediction variance, and the examples include the bagging methods and  forests of randomized trees \citep{Breiman96Bagging,Breiman01RandomForests}.

In order to make use of the assumption that these discriminative voxels are often spatially contiguous and result in distributed clusters,
one  proposed  the idea of using common stability selection together with clustering \citep{Gramfort12RandomizedSparsityClustering,Varoquaux12Clustering}. Specifically, the clustering will be run after subsampling on training samples and random rescaling of features during each resampling of stability selection. The added clustering helps to improve the conditioning of
resulted sub-matrices of the training data matrix. However,  
%
 the random ``rescaling"  during their 
   implemented stability selection 
 is voxel-wise and fails to consider the spatial contiguity of the clustered discriminative voxels. 
\subsection{Our focus and  contributions}

In this paper, we propose a variant of stability selection based on structural sparsity, called ``randomized structural sparsity". It is   implemented via the adoption of  the ``constrained  block subsampling" technique for voxel-wise fMRI data analysis, in contrast to  single voxel-wise subsampling in the classical stability selection. We expect it to achieve an improved sensitivity of the selected discriminative voxels. 
We   show empirically that 
%
this ``blocked" variant of stability selection can achieve significantly better sensitivity than  alternatives, including the original stability selection,  while keeping the control of false positives for voxel selection. 
%


We need to point out that this new algorithm is  beyond a simple summation of stability selection and structural stability. It has the following extra important  advantage: in many cases where  structural information such as clustering structures is only a rough approximation, i.e.  neighboring voxels in the same brain area might be highly correlated though not necessarily all informative, a.k.a. discriminative,  the subsampling scheme can help remedy this via supervised refining and  outlining of the true shapes of the discriminative regions, as showed by numerical experiments. Compared with Randomized Ward Logistic algorithm proposed in \citep{Gramfort12RandomizedSparsityClustering}, our algorithm only needs to perform clustering once, and therefore is computationally more efficient. 

The rest of the paper is organized as follows. In section \ref{Sec:Method}, we introduce our new algorithm for stable voxel selection. In section \ref{Sec:Experiments}, we demonstrate the advantages of our algorithm based on both synthetic data and real fMRI data in terms of  higher sensitivity and specificity. In section \ref{Sec:conclusion}, a short summary of our work and  possible future research directions will be given. 

\section{The Proposed Method} \label{Sec:Method}

\subsection{Background and Motivation }


%
  Let us denote an fMRI data matrix as $\mathbf{X} \in \mathbb{R}^{n\times p}$ where $n$ is the number of samples and $p$ is the number of voxels with $n \ll p$, and corresponding classification information as $\mathbf{y} \in \mathbb{R}^{n\times 1}$. Here we  consider only the binary classification and $\mathbf{y}_i \in \{1,-1\}.$ 
While our main ideas can be applied to  other models, we take the following sparse logistic regression for classification as our example  to show the existing difficulties and our corresponding efforts, in detail. 
\begin{eqnarray}\label{eq:L1}
\min_{\mathbf{w}}\|\mathbf{w}\|_1 + \lambda \sum_{i=1}^{n} \log(1+\exp(-\mathbf{y}_i(\mathbf{X}_i^{T} \mathbf{w}+c)))
\end{eqnarray}
where $\mathbf{X}_i$ denotes the $i$-th row of $\mathbf{X}\in\mathbb{R}^{n\times p}$; $\mathbf{w} \in \mathbb{R}^{p\times 1}$ is the weight vector for the voxels and $c$ is the intercept (scalar). The voxels corresponding to $\mathbf{w}_i$ with large absolute value will be considered as the discriminative voxels. 

Structured sparsity models beyond the plain $\ell_1$ norm regularized models  have
been proposed to  enforce  more structured
constraints on the solution \citep{Bach12StructuredSparsity,Li13ClusterGuided,Mairal13SFS}, where the structure can be defined based on the feature correlation. As an important special case,   
the common way to make use of the clustering or grouping structure is to adopt the group sparsity induced norm \citep{Bach12OSP}, as follows.
\begin{eqnarray}\label{eq:structuralL1}
\min_{\mathbf{w}}\sum_{g\in \mathcal{G}}\|\mathbf{w}_g\|_2 +  \lambda \sum_{i=1}^{n} \log(1+\exp(-\mathbf{y}_i(\mathbf{w}^{T}\mathbf{X}_i+c))),
\end{eqnarray}
where $\mathcal{G}$ is the grouping information. Compared with (\ref{eq:L1}), the main difference is the regularization term;  we are using a mixed $\ell_1/\ell_2$ norm.  The model (\ref{eq:structuralL1}) belongs to the family of structural sparsity regularized feature selection models.  
The resulting penalty incorporating the parcellation information has been shown to improve the prediction performance and interpretability of the learned models, provided that the grouping structure is relevant \citep{Yuan06groups,Huang2010,jenatton12multiscale,Bach12StructuredSparsity}. In addition, the number of selected candidate
features is allowed to be much larger when an additional group structure is incorporated, particularly
when each group contains considerable redundant features \citep{Jenatton11StructuredSparsity,Xiang2015NonconvexGroup}.  Therefore, the parcellation is able to help improve the sensitivity of voxel selection \citep{Flandin02agglomeration}.

However,  the group sparsity-induced norm regularized model (\ref{eq:structuralL1}) is expected to improve the sensitivity  with respect to the plain $\ell_1$ norm regularized model (\ref{eq:L1}) due to the adopted mixed $\ell_{2,1}$ norm only if the grouping information $\mathcal{G}$ is reliable enough. Obtaining  an appropriate $\mathcal{G}$ might be possible in practice 
 from either the prior anatomical knowledge or data-driven methods based on the voxel correlation.
  However, many methods of obtaining  $\mathcal{G}$  are not incorporating the available classification or labelling information. Therefore,  it is possible that  only a subset of voxels  in a certain group is discriminative. In such case, the model (\ref{eq:structuralL1}) often fails to make a segmentation, because it is likely to simultaneously choose all the voxels of a certain group or  simultaneously choose none of them, due to the adoption of the $\ell_2$ norm.   In addition, just like the plain $\ell_1$ norm regularized model, the difficulties of  choosing  a proper regularization parameter and lack of finite sample control of false positives still exist.  
As mentioned above, an effective way to control the false positives and reduce the difficulty of choosing the proper regularization parameter when applying the sparsity regularization based models is stability selection \citep{Meinshausen10StabilitySelection}, which has been applied for voxel selection or connection  selection in brain image analysis \citep{Rondina14,Ye12SparseLearningSS,Cao2014Sparse,Ryali20123852}. 

However, while the control of false positives can be achieved,  a large false negative rate is often expected, especially in the case of redundant and correlated voxels, because this correlation prior is not explicitly taken into consideration.  

\subsection{The Key Component: Randomized Structural Sparsity} \label{Sec:ConstrainedBSub}
In this paper, we aim to stably identify  the discriminative voxels including those probably correlated ones,  for better interpretation of  discovered potential biomarkers. To achieve this goal, we incorporate the spatial structural knowledge of voxels into the stability selection framework. The novelty of our research is to propose a ``\textit{randomized structural sparsity}", which aims to integrate the stability selection and the common ``structural sparsity".  

One important component of ``randomized structural sparsity" is the subsampling based stability selection \citep{Beinrucker12ExtensionSS}, rather than the original reweighting-based stability selection \citep{Meinshausen10StabilitySelection}.  It has been shown that the former is likely to yield an
improvement over the latter whenever the latter itself improves over
a standalone pure $\ell_1$ regularization model \citep{Beinrucker12ExtensionSS}. Moreover, subsampling  is easier to  extend to block subsampling and  combine with structural sparsity. 

 Let us first briefly explain  subsampling-based stability. For the training data matrix $\mathbf{X}\in\mathbb{R}^{n\times p}$, subsampling based stability selection consists of applying
the baseline, i.e. the pure $\ell_1$ regularization model such as (\ref{eq:L1}), to random submatrices of $\mathbf{X}$ of size $[n/L] \times [p/V]$, where $[]$ is  rounded off to the nearest integer number, and returning those features having the largest selection frequency. The original stability selection \citep{Meinshausen10StabilitySelection} can be roughly considered as a special case, where $L=2$ and $V=1$, except that the original stability selection \citep{Meinshausen10StabilitySelection} reweighs each feature (voxel, here) by a random weight uniformaly sampled in $[\alpha,1]$ where $\alpha$ is a positive number, and subsampling can be intuitively seen as a crude version of this
by simply dropping out randomly a large part of the
features \citep{Beinrucker12ExtensionSS}. 
The other important component of randomized structural sparsity
is to incorporate  structural information, such as the parcelling information of  the brain into consideration. The kind of partition information is based on either the prior anatomical knowledge of brain partition \citep{TzourioMazoyer2002AnatomicalParcellation}, or the clustering results based on the fMRI data, as done in the structural sparsity model (\ref{eq:structuralL1}).

The above ``randomized structural sparsity" is a general concept and might have different specific implementations in practice, depending on different data types and applications. For  voxel-wise fMRI data analysis, we propose a specific implementation named  ``\textit{constrained block subsampling}'', where by ``constrained" we mean that the parcelling information will be respected to certain degree. The block subsampling \citep{Lahiri99Blockbootstrap} is adopted because it is able to
replicate the correlation  by subsampling  of blocks of data.  

Specifically,  each cluster $g \in \mathcal{G}$,  consists of highly correlated voxels. 
%
%
%
After the  block subsampling, 
the selected voxels from the same cluster will be considered as a group. 
In particular, 
 the chosen voxels lying in a  cluster $g \in \mathcal{G}$ are noted as  a set $g' \subseteq g$. In addition, in order to make  every brain partition, especially those of small sizes  have a chance to be sampled during the block subsampling, we borrow some idea of ``proportionate stratified sampling" \citep{Sarndal03stratified,Devries86}, i.e. the same sampling fraction is used within each partition. The purpose is to reduce the false negatives, especially when the sizes of different partitions are of quite a range. 
%
%
Correspondingly, one can solve the following group-sparsity based recovery model. 
\begin{eqnarray}\label{eq:structuralL1Subsampling}
\min_{\mathbf{w}'}\sum_{g' \subset g\in \mathcal{G}}\|\mathbf{w'}_{g'}\|_2 +  \lambda \sum_{i \in \mathcal{J}} \log(1+\exp(-\mathbf{y}_i(\mathbf{w'}^{T}\mathbf{X}'_i+c)))
\end{eqnarray}
where $\mathbf{w}'$ and $\mathbf{X'}$ are corresponding parts of $\mathbf{w}$ and $\mathbf{X}$, respectively,  based on the selected voxels during the subsampling, and  $\mathcal{G}$ is a predefined partitions of the brain, based on the either biological knowledge or data driven learning or estimation such as clustering. $\mathcal{J}$ is the set of the indices of the selected samples during the current subsampling.

Notice that while  ``constrained block subsamplings"  respects the prior knowledge $\mathcal{G}$, it also provides the flexibility that the resulting discriminative regions can be of any shape, and the final selected voxels of each cluster can be only a portion of all of it, because  the subsampling  makes the selection frequency score be
 able to outline shapes of the true discriminative regions, whose sizes may not be exactly the same as the sizes of the original partitions defined by $\mathcal{G}$.  This kind of flexibility is important because 
the neighboring voxels   belonging to the same
brain area are not necessarily all significantly discriminative voxels, though they might be  highly correlated.  In other words, we aim to seek sets of correlated voxels with
similar associations with the response (or labels), if only part of but not all of the correlated features have a similar
association with the response, as mentioned in  \citep{Witten14clusterElasticNet}.

Furthermore, for our case of small samples and very high dimensional feature space, we need to consider the bias-variance dilemma or bias-variance tradeoff  \citep{Geman:1992:NNB}. In general, we would like to pay a little bias to save a lot of variance, and dimensionality reduction can decrease variance by simplifying models \citep{James13StasiticalLearning}.  Correspondingly, while we can still use the (\ref{eq:structuralL1Subsampling}) as the baseline subproblem for our stability selection framework,
we prefer a simple  ``averaging" idea \citep{Varoquaux12Clustering} applied to  (\ref{eq:structuralL1Subsampling}), 
because \citep{Park01042007} has showed that when the variables or features were positively correlated, their average was a strong feature, and this yielded a fit with lower variance than the individual variables. Specifically,
by averaging the voxels picked by the block subsampling lying in the same group as a single super-voxel,  the model  (\ref{eq:structuralL1Subsampling}),   can be further reduced to the following low dimensional version
\begin{eqnarray}\label{eq:everaged}
\min_{\tilde{\mathbf{w}}}\sum_{g' \subset g\in \mathcal{G}}|\tilde{\mathbf{w}}_{g'}| +  \lambda \sum_{i \in \mathcal{J}} \log(1+\exp(-y_i(\tilde{\mathbf{w}}^{T}\mathbf{\tilde{X}}_i+c)))
\end{eqnarray}
where $\tilde{\mathbf{w}} \in \mathbb{R}^q$, and $q$ is the number of clusters. $\tilde{\mathbf{w}}_{g'} $ is an average of voxels in the subset $g'$ of cluster $g\in \mathcal{G}$, and $\mathbf{\tilde{X}}\in \mathbb{R}^{[\alpha n] \times p}$ is the corresponding averaged $\mathbf{X}$.
 Thus the number of variables in the sparse recovery model (\ref{eq:everaged}) is greatly reduced to the number of clusters. This way, the resulted  recovery problem  (\ref{eq:everaged}) is of much smaller scale and therefore can be solved quite efficiently. In addition,  the properties of the resulting new data matrix $\tilde{\mathbf{X}}$ is greatly improved due to de-correlation via the clustering of correlated columns. The analysis of a better-posed compatibility constant for the $\tilde{\mathbf{X}}$ was proposed in  \citep{Buhlmann2013correlated}.   The idea of averaging, also called feature agglomeration \citep{Flandin02agglomeration}, was also applied in \citep{Gramfort12RandomizedSparsityClustering}.  
%
%
If the $j$-th column of $\tilde{\mathbf{X}}$ is selected due to the large magnitude of $\tilde{\mathbf{w}}_j$, then its represented picked blocked voxels lying in the group $g^{(j)} \in \mathcal{G}$ ($j=1,2,\ldots, q$) of $\mathbf{X}$ are all counted to be selected, in 
the non-clustered
space.  Its corresponding score $\mathbf{s}_i$ will be updated ($i=1,2,\ldots,p$). Notice that the averaging of subsamplings is more than a simple spatial smoothing, due to different sumsampling results of different stability selection iterations. Therefore, the boundaries of the detected discriminative regions can be trusted to certain accuracy. 

\subsection{Algorithmic framework} \label{Sec:AlgIntro}


We first obtain the structural information about the brain. Here we perform a data-driven clustering operation to partition the voxels into many patches according to their strong local correlations. 
In our algorithm, both the common K-means and the spatially constrained spectral clustering algorithm \citep{Craddock13connectomes} implemented written as a Python software (\url{http://www.nitrc.org/projects/cluster_roi/}) are used in our experiments.  We denote the set of the groups via the clustering algorithm as $\mathcal{G}$, whose cardinality is denoted as $q$, which is usually much less than $p$ and comparable with $n$.  Notice that in (\ref{eq:everaged}), the number of unknowns is reduced from $p$ to the number of clusters, i.e. $q$. While the number of samples is a fraction of the total samples, for example, $[n/2]$. In this paper, we typically choose the $q$ at least $2$ times larger than
the number of samples but smaller than $5$ times of the number of samples in practice.  

Next comes the ``constrained block subsamplings".  Denote the number of resamplings as $K$.
This blocked variant of stability selection is different from the classical stability selection in terms of the subsampling on the features, i.e. the columns of the data  matrix $\mathbf{X}$. But it shares the same way as the classical stability selection when performing subsampling of the observations, i.e. the rows of the data  matrix $\mathbf{X}$.
Let the subsampling fraction be $\alpha \in [0,1]$ and let $\mathcal{J}$ denote the indices of selected rows   and the cardinality of $\mathcal{J}$ is $[\alpha n]$, where $[]$ is   rounded off to the nearest integer number.
%
Then ``contrained block subsamplings" are applied to the voxels, i.e. the columns of  $\mathbf{X}$ as mentioned in last section.  Notice that our algorithm only runs the clustering once and the following ``constrained block subsamplings" resulted in a much smaller size of $\ell_1$ problem, where the number of unknows is equal to the number of clusters. Therefore, our algorithm is not computationally expensive.  

The procedure of our algorithm is summarized below. 

\ni \hrulefill

\ni {\bf The Algorithmic Framework of Constrained Blocked Stability Selection Method:}\\

\ni {\bf Inputs:}
\begin{itemize}
\item[(1)] Datasets $\mathbf{X} \in \mathbb{R}^{n\times p}$

\item[(2)] Label or classification information $\mathbf{y} \in \mathbb{R}^n$

\item[(3)] Sparse penalization parameter $\lambda>0$

\item[(4)] Number of randomizations $K$ for each stage;   subsampling fraction $\alpha \in [0,1]$ in terms of rows of $\mathbf{X}$; subsampling fraction $\beta \in [0,1]$ in terms of columns of $\mathbf{X}$;

\item[(5)] Initialized stability scores:  $\mathbf{s}_{i}=0.$ $(i=1, 2, \ldots, p)$
\end{itemize}
\ni {\bf Output:} Stability scores $\mathbf{s}_{i}$ for each voxel.  $(i=1, 2, \ldots, p)$
\vspace{2mm}





 Obtain a brain parcellation. For example, perform the clustering of voxels based on their spatial correlation and denote the number of clusters as $q$\\

 {\bf for} k=1 to K\\\vspace{0.4cm}

\quad 1:  \quad Perform  sub-sampling in terms of rows: $X \leftarrow  X_{[J,:]}, y \leftarrow  y_{\mathcal{J}}$ where $\mathcal{J}\subset\{1, 2, \ldots, n\},$ $\text{card}(\mathcal{J})=[\alpha n]$, the updated $X\in R^{[\alpha n]\times p}$, and the updated $\mathbf{y}\in R^{[\alpha n]}$. \vspace{0.5cm}

 \quad 2:  \quad Perform constrained block subsampling in terms of columns (voxels):  $X' \leftarrow X_{[:,\mathcal{I}]}$, where $\mathcal{I}\subset\{1, 2, \ldots, p\},$ and $\text{card}(\mathcal{I})=[\beta p]$ \vspace{0.5cm}

 \quad 3:  \quad Use the current clustering, and calculate the mean of randomly picked voxels within each cluster: $\tilde{X} \leftarrow \text{mean}(X'), \tilde{X} \in R^{\alpha n\times q}$\vspace{0.5cm}

\quad 4:  \quad Estimate $\tilde{\mathbf{w}} \in \mathbb{R}^q$ from $\tilde{X}$ and $\mathbf{y}$ with sparse logistic regression (\ref{eq:everaged}).\vspace{0.5cm}

 \quad 5:  \quad Set weights for the randomly picked voxels with estimated coefficients of the averaged voxels: $\mathbf{w}^{(k)} \leftarrow \tilde{\mathbf{w}}$,
$\mathbf{w}^{(k)} \in \mathbb{R}^{[\beta p]}$\vspace{0.5cm}

 \quad 6: $\mathbf{s}_i=\mathbf{s}_i +1$, if $i \in \text{supp}(\mathbf{w}^{(k)})$, for $i=1, 2, \ldots, p.$\vspace{0.5cm}

 {\bf end for}



\ni\hrulefill

\subsection{Some Preliminary Rethinking of Our Algorithms}
Basically, the original stability selection proposed in \citep{Meinshausen10StabilitySelection} is mainly on random subsampling of observations, i.e. the rows of $\mathbf{X}$. As the paper by \citep{Beinrucker15sssCovariates} has also pointed out,  the random subsampling in terms of observations  can in general guarantee the finite control of false positives, even though different base methods are adopted. Therefore, while we are using a more complicated base method  (\ref{eq:everaged}) than the plain $\ell_1$ norm regularized model, the finite control of false positives  can be still achieved. However, the corresponding new theoretical result in terms of bounding the ratio of the expected number of false positive selections over the total number of features (false postive rate) needs to be addressed in the future work.

It is natural that we adopt the structural sparsity regularized models such as (\ref{eq:everaged}), as the base methods of stability selection. As \citep{Bach12StructuredSparsity,Bach12OSP,Flandin02agglomeration} pointed out, the regularization term incorporating the  parcellation information has been shown to improve the interpretability of the learned models and  the detection sensitivity of voxel selection for the functional MRI data, provided that the parcellation information is quite relevant.

However, the parcellation information might be not very accurate.  Any fixed brain parcellation indeed might bring certain degree of bias or arbitrariness. 
In this paper, we turn to help of the block subsamplings. While we present some intuitive explanation in Section \ref{Sec:ConstrainedBSub},  a thorough study of the effect of block subsampling on reducing the arbitrariness is not presented in this paper and constitutes an important future research topic.

Here we need to point out that the method proposed in \citep{Varoquaux12Clustering} does not suffers the bias caused by a fixed parcellatoin, because the clustering is performed on each step of stability selection after the randomized rescaling on each feature.
However, from the computational point of view, our adopted onetime parcellation helps improve the computational efficiency, because clustering takes a large proportion of running times of both our algorithm and the algorithm by \citep{Varoquaux12Clustering}. Some preliminary comparison of running time of different algorithms are presented in Section \ref{runningtime}. 

\section{Numerical Experiments} \label{Sec:Experiments}
In this paper, we  compare our algorithm with the classical univariate voxel selection method, and with
 state-of-the art multi-voxel pattern recognition methods, including  T-test, $\ell_2$-SVM, $\ell_2$ Logistic Regression,  $\ell_1$-SVM, $\ell_1$ Logistic Regression,  randomized  $\ell_1$ logistic regression,  Smooth Lasso \citep{hebiri2011} and TV-L1 \citep{Gramfort:2013:IPR:2552484.2552506} and Randomized Ward Logistic \citep{Varoquaux12Clustering}.  Here randomized  $\ell_1$ logistic regression is based on the original stability selection \citep{Meinshausen10StabilitySelection} and random reweighing on the features.

The T-test is implemented as an internal function in MATLAB. $\ell_2$-SVM, $\ell_2$ Logistic Regression,  $\ell_1$-SVM, and $\ell_1$ Logistic Regression, have been implemented in LIBLINEAR \citep{REF08a} or SLEP (Sparse Learning with Efficient Projections) software \citep{Liu09:SLEP:manual}. Randomized  $\ell_1$ logistic regression is written based on the available  $\ell_1$ logistic regression code.
TV-L1 and Randomized Ward Logistic are implemented in Python, and integrated into NiLearn, a great Python software for NeuroImaging analysis \url{http://nilearn.github.io/index.html}. We were kindly provided with the source code by their developers.  For the hyper-parameters such as the regularization parameters, their choices are mostly based on cross validation unless specified otherwise. 

\subsection{Settings of Algorithms}
 For our algorithm, 
the block size might affect the performance of our algorithm \citep{Lahiri01BlockSize}.  Given the number of blocks,
there is an inherent trade-off in the choice of  block size.
When only a very limited number of randomizations are allowed, big blocks will most likely not match the geometry
of the true support and easily result in many false positives. But the condition of too small blocks is likely to result in many false negatives due to the likely ignorance of the local correlations of neighboring voxels.
 The block size is not optimized in the following experiments via the probable prior knowledge of the discriminative regions, but it still achieves an impressive performance. It was set to be $3\times 3$ in synthetic data and $4 \times 4 \times 4$ in the real fMRI data experiment, respectively. We set the subsampling rate $\alpha=0.5$ and $\beta=0.1$. For our synthetic data and and  the real fMRI data, we set the total resamplings  $K=50$ and $K=200$, respectively. The resampling number of random $\ell_1$ logistic regression is $500$ in all of our experiments. The choice of the number of resamplings is only empirical here. 



 \subsection{Evaluation Criteria}

We would like to  demonstrate that our method can achieve  better control of  false positives and false negatives than alternative methods, due to our  incorporation of  both stability selection and structural sparsity.

  For the synthetic data, we can directly use the precision-recall curve since we know the true discriminative features. Precision (also called positive predictive value) is the fraction of retrieved instances that are relevant, while recall (also known as sensitivity) is the fraction of relevant instances that are retrieved.
  We also plot the first $T$ discriminative voxels discovered by different algorithms, where $T$ is the number of true discriminative features. It provides  a snapshot to display how many noisy features are included in the selected features of different algorithms.

For real fMRI data, we first show brain maps obtained by the feature weights, which are
not thresholded for visualization purposes, meaning
that the zeros obtained are actually zeros. We are able to determine  the discriminative regions, revealed by different algorithms, based on our visual inspection. 

In addition to vision inspection and experience, we  would also like to find an objective threshold. In general, voxels whose corresponding weights have larger magnitude than this threshold will be considered as the discriminative voxels and will be shown in the brain maps. 
However, the setting of a threshold value is quite difficult and may adopt different schemes in different situations. While a through study of threshold setting is out of reach of this paper, we consider to 
use the cross-validation based on prediction accuracy and find out the threshold value corresponding to the highest prediction accuracy. However,  as \citep{hofner2014controlling} has pointed out that the prediction accuracy and variable selection are two different goals.  Different features might result into the same level of classification accuracy. Therefore, it is often acceptable to develop some heuristic for setting the threshold value for feature selection, beyond cross validation, such as the method by \citep{fellinghauer2013stable}. Another kind of important method to set the threshod value is based on FDR control with multivariate p-values \citep{chi2008false}. 


The main feature of our algorithm is its improved sensitivity while maintaining good specificity. In order to prove that  the extra probably discriminative regions discovered  only by our algorithm are  true and stable positives, we adopt the following two methods. One is to take these extra selected voxels to construct a classifier to perform classification on the test data. A satisfying classification accuracy can at least prove the existence of true positiveness. Notice that, while we have mentioned that the prediction accuracy is not very reliable criterion for model selection,  high prediction accuracy can in generally tell us that at least portion of these voxels are truly discriminative. 
If a high prediction accuracy is achieved,  we are likely to believe that the corresponding  brain regions are discriminative, though their sizes might be not accurate.   The other is to perform a false positive estimation scheme based on a permutation test in order to calculate the ratio of false positives among all the finally selected voxels \citep{Rondina14}.

\subsection{Synthetic Data}
We simulated simple case control analysis model and work on $46\times 55 \times 46$ brain images including $27884$ voxels of interest. We generated $50$ observations for each group, i.e. the control group and the case (patient) group.

   \begin{figure}[!h]
    \centering
    \includegraphics[width=0.99\textwidth,height=0.55\textwidth]{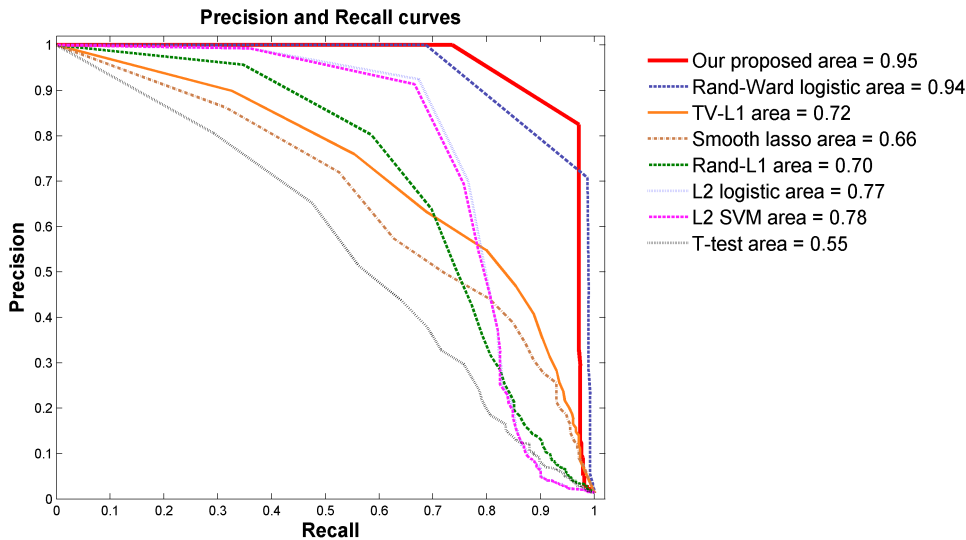} 
    \vspace{-0.1cm}\caption{Precision-Recall Curve on the new Synthetic Data:  Our method can achieve the best control of both false positives and false negatives, as well as the largest AUC (Area Under  Curve) value .
   }\label{Fig:Precision_recall}
\end{figure}

\begin{figure}[!h]
    \centering
    \includegraphics[width=0.95\textwidth,height=0.85\textwidth]{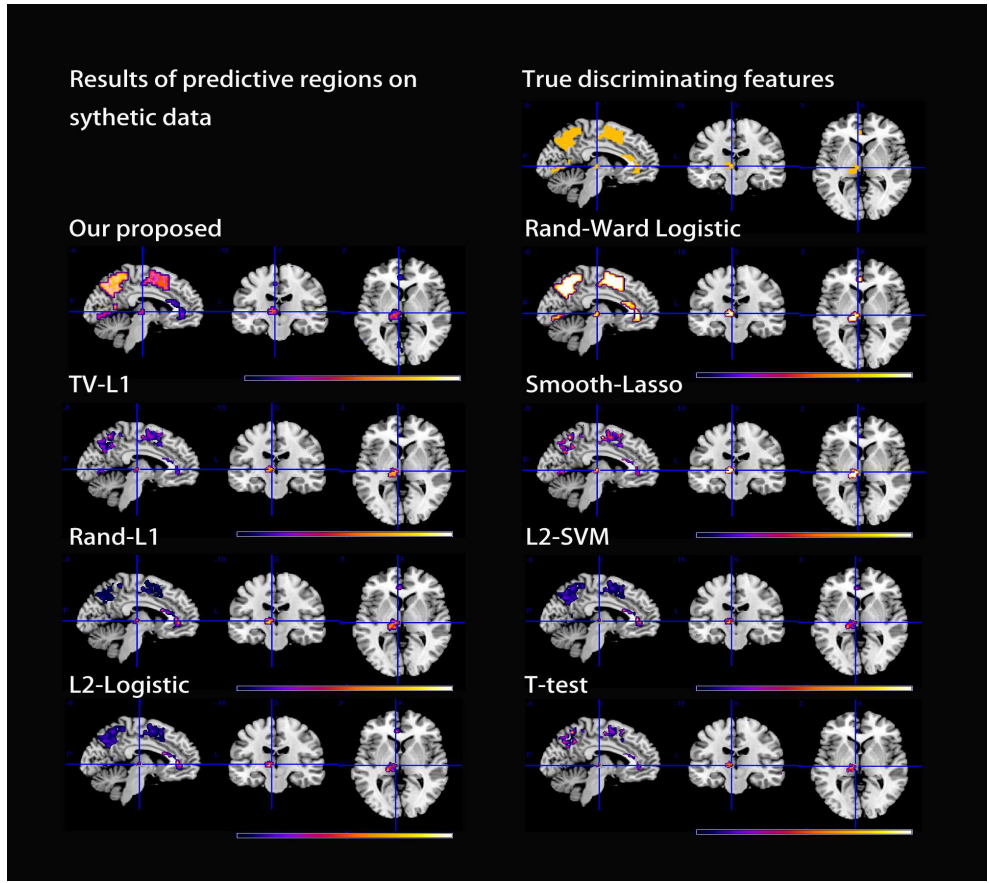}
    \vspace{-0.1cm}\caption{Result on the  Synthetic Data: Our algoritm together with Randomized Ward Logistic algorithm can find more true discriminative voxels with very few noisy voxels. 
   }\label{Fig:Synetic_region}
\end{figure}

There were five discriminated clustered features with $383$ total voxels in the frontal lobe, parietal lobe, occipital lobe, and subcortical regions. Each  cluster contained more than $30$ voxels, as showed in Figure \ref{Fig:Synetic_region} in yellow.

The elements in the first two clustered features: $x(k)_{i,n} = k+\epsilon(k)_{i,n}$ (case), $y(k)_{j,n} = \eta(k)_{j,n}$ (control), where $i, j = 1, 2, . . . , 50$ representing the index of persons of each group, and $k = 1, 2$ representing the index of the first two clustered features, and $n = 1, 2, . . . , 100 $ representing the index of features of each cluster. $\epsilon(k)_{i,n}$ and $\eta(k)_{j,n}$ are Gaussian i.i.d distributed. The elements in the other three clustered features are spatially distributed patterns, which are Gaussian i.i.d distributed and constrained by linear condition: $x(k)_{i,n} = \epsilon(k)_{i,n},$ $x(k)_{j,n} = \eta(k)_{j,n}$  $\sum_{k=3}^5 x(k)_{i,n} > 1$ (case), and $\sum_{k=3}^5 y(k)_{j,n} < 1$ (control), where $i, j = 1, 2, . . . , 50$ representing the index of persons of each group, and $k = 3, 4, 5$ representing the index of the last three clustered features, and $n = 1, 2, . . . , 100$ representing the index of features of each cluster. As above, $\epsilon(k)_{i,n}$ and $\eta(k)_{j,n}$ are also Gaussian i.i.d distributed. The features were spatially clustered in different brain regions. We also simulated the other voxels in whole brain image X as Gaussian noise. Notice that these are distributed multivariate discriminative patterns, each of which consists of $3$ voxels from each of the last $3$ clusters, respectively.   For the clustering algorithm used in our algorithm, we use K-means and the number of clustering is equal to $200.$



We would like to show that our method can achieve both  accuracy and completeness in terms of discovery of  discriminative features. Here accuracy means a small false positive rate and completeness means a small false negative rate.
In Figure \ref{Fig:Precision_recall}, we use Precision-Recall Curve to demonstrate this advantage of our method. Precision (also called positive predictive value) is the fraction of retrieved instances that are relevant, while recall (also known as sensitivity) is the fraction of relevant instances that are retrieved. While still keeping good control of false positives, our algorithm, together with Randomized Ward Logistic are  the most sensitive, i.e. discovering the almost ``complete" set of discriminative features. Notice that the standard stability selection, i.e. randomized $\ell_1$ algorithm does not work well in this case.

In Figure  \ref{Fig:Synetic_region}, we plot out the identified discriminative voxels corresponding to the top $383$ weights of largest magnitude of these different involved algorithms, where $383$ is the number of true discriminative voxels displayed in the subplot of the most upper-right corner. Our algorithm, together with the Randomized Ward Logistic algorithm, discover more clustered true positive features than others. Moreover, our algorithm is computationally more efficient than Randomized Ward Logistic because we  run the clustering only once. Notice that the synthetic data is 3-D and therefore is hard to  visualize in  2-D. So the performance comparison of each algorithm is more directly displaying via the Precision-Recall curve of Figure \ref{Fig:Precision_recall}. Nevertheless, Figure  \ref{Fig:Synetic_region} can be a useful supplement, as an illustration of the performance of the achieved sensitivity and specificity of different voxel-selection algorithms.

\subsection{Real fMRI Data- Chess-Master Data}

In this experiment, we aim to identify the brain activation pattern of a Chinese-chess  problem-solving task in  professional Chinese-chess grandmasters. 
$14$ masters on Chinese chess were recruited and studied.
All subjects were right-handed and had no history of psychiatric or neurological disorder. 
During the fMRI scanning, subjects were presented with two kinds of stimuli: a blank chessboard and patterns of Chinese chess spot game with checkmate problems. Each condition was presented for 20s, with a 2s-long break between. The block was repeated nine times with different problems in each block. The break between each block is also 2s. There were $9$ blocks overall.
In consideration of the delay of Blood Oxygenation Level Dependent (BOLD) effect \citep{aguirre1998variability} and the hypothesis that the master may solve the problem in less than 20s, we  selected the 4th-8th images of each state in each block for classification. That is, the number of observations of each subject for classification is $90$, among which $45$ are in blank states while the other $45$ are in task states. We used an averaged data from all  $14$ grandmasters.
  Data Acquisition and Preprocessing
Scanning was performed on a 3T Siemens Trio system at the MR Research Center of West China Hospital of Sichuan University, Chengdu, China. T2-weighted fMRI images were obtained via a gradient-echo echo-planar pulse sequence (TR, 2000ms; TE, 30ms; flip angle=$90^\circ$; whole head; 30 axial slice, each 5mm sick (without gap); voxel size=$3.75\times 3.75\times 5mm^3$).
fMRI images were preprocessed using Statistical Parametric Mapping-8 (SPM8, Welcome Trust Centre for Neuroimaging, London, UK. http://www.fil.ion.ucl.ac.uk/spm). Spatial transformation, which included realignment and normalization, was performed using three-dimensional rigid body registration  for head motion. The realigned images were spatially normalized into a standard stereotaxic space at $2 \times 2 \times 2$ $mm^3$, using the Montreal Neurological Institute (MNI) echo-planar imaging (EPI) template. A spatial smoothing filter was employed for each brain¡¯s three-dimensional volume by convolution with an isotropic Gaussian kernel (FWHM= 8 mm) to increase the MR signal-to-noise ratio. Then, for the fMRI time series of the task condition, a high-pass filter with a cut-off of 1/128 Hz was used to remove low-frequency noise. Among all $90$ fMRI samples, each of them  was of size $91\times 109\times91$.  For the clustering algorithm used in our algorithm, we use K-means and the number of clusters is equal to $200.$

\begin{figure}[!h] 
    \centering
    \includegraphics[width=0.95\textwidth,height=0.7\textwidth]{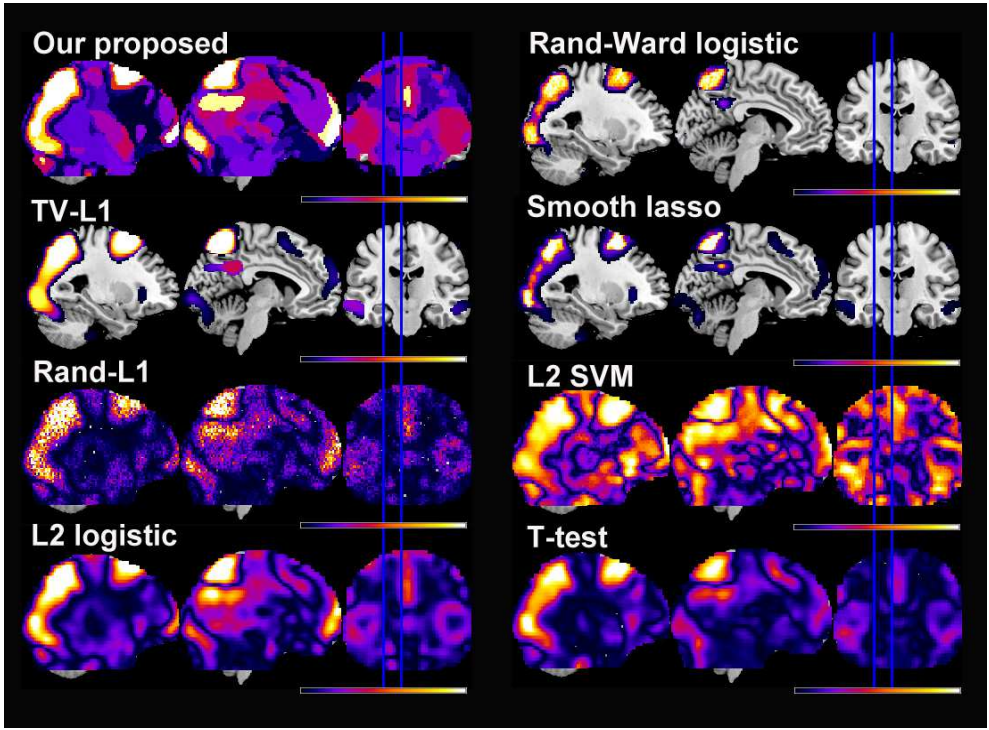}
    \vspace{-0.1cm}\caption{Score maps (unthresholded)  estimated by different methods. The most significant areas discovered by our method are quite spatially contiguous and are of high contrast with other areas.
   }\label{Fig:chess_unthreshold}
\end{figure}


\begin{figure}[!h]
    \centering
    \includegraphics[width=0.95\textwidth,height=0.7\textwidth]{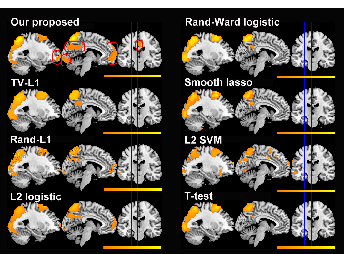}
    \vspace{-0.1cm}\caption{Score maps (thresholded)  estimated by different methods. Our algorithm reveals more predictive areas.
   }\label{Fig:chess_threshold}
\end{figure}


Figure \ref{Fig:chess_unthreshold} shows brain maps based on the weights or scores of voxels of different algorithms.
The scores are
not thresholded for visualization purposes, meaning
that the zeros obtained are actually zeros. One can observe that despite being fairly
noisy,  the most significant localized discriminative regions of the
brain, identified by different algorithms, can be visually recognized. Even with the same number of selected voxels, our method
is expected to achieve the best balance of controlling of both the false positives and false negatives.

In general, when identifying the potential biomarkers, controlling the false positives should be the first priority. They need to be treated carefully and  controlled strictly. So we need to  set a threshold value to filter out at least the apparent noisy features, which are either too scattered or  in the wrong regions from the existing confirmed knowledge. We carefully  set the threshold values for results of different algorithms in order to  control the false positives and obtain a  cleaner brain map. Notice that in this experiments, a common threshold-setting method based on the prediction accuracy via cross validation does not work well, because this is a very simple cognitive task and the involvement of many noisy features or using only a small number of true positives can also achieve nearly 100\% accuracy. 
Figure \ref{Fig:chess_threshold} is the result after thresholding out the apparent noisy features as either too scattered or in the unreasonable area.
%
%
The false positives are expected to be well controlled. We can see that our algorithm is most sensitive and identified several extra brain regions. We construct  classifiers based on each of these extra regions and test their predictive power. They are more than 95\%  accurate, so these extra regions are very likely to be relevant.

We take a further look at the result of Figure \ref{Fig:chess_threshold} from the viewpoint of brain science. All the multivariate pattern feature selection methods successfully identified at least partial task-related prefrontal and parietal and occipital lobe regions. These results indicate a co-working pattern of the cognitive network and default mode network of the human brain during our board game task state. However, compared with most alternative algorithms besides the common stability selection, our proposed method identifies much more brain regions in the medial prefrontal cortex and precuneus gyrus that are functional and structural central hubs in the default mode network and in occipital lobe which contains parts of visual cortex. The common stability selection, i.e. randomized $\ell_1$ logistic regression is able to identify the medial prefrontal gyrus, but it misses the precuneus. Moreover, the common stability selection is likely to return a result that is  slightly more scattered, which does not match the second hypothesis about continuousness and compactness. Of even greater concern is that fact that its scattered results make it difficult to distinguish between true positives from false positives. In addition, common stability selection required many more subsamplings, for  example, $500$ times here compared with  our method which only takes $50$ subsamplings. This result verifies one of the main advantages of our method, namely its computational efficiency, which is especially important for  high dimensional problems. In section \ref{runningtime}, we will present the running time comparison of different algorithms.  Our method also has  better inference quality due to the incorporation of  prior structural information of the fMRI data. As mentioned before, this computational efficiency also comes from the even smaller size of the subproblem \eqref{eq:everaged} due to the adoption of the averaging idea within a cluster.



\begin{figure}[!h]
    \centering
    \includegraphics[width=0.95\textwidth,height=0.85\textwidth]{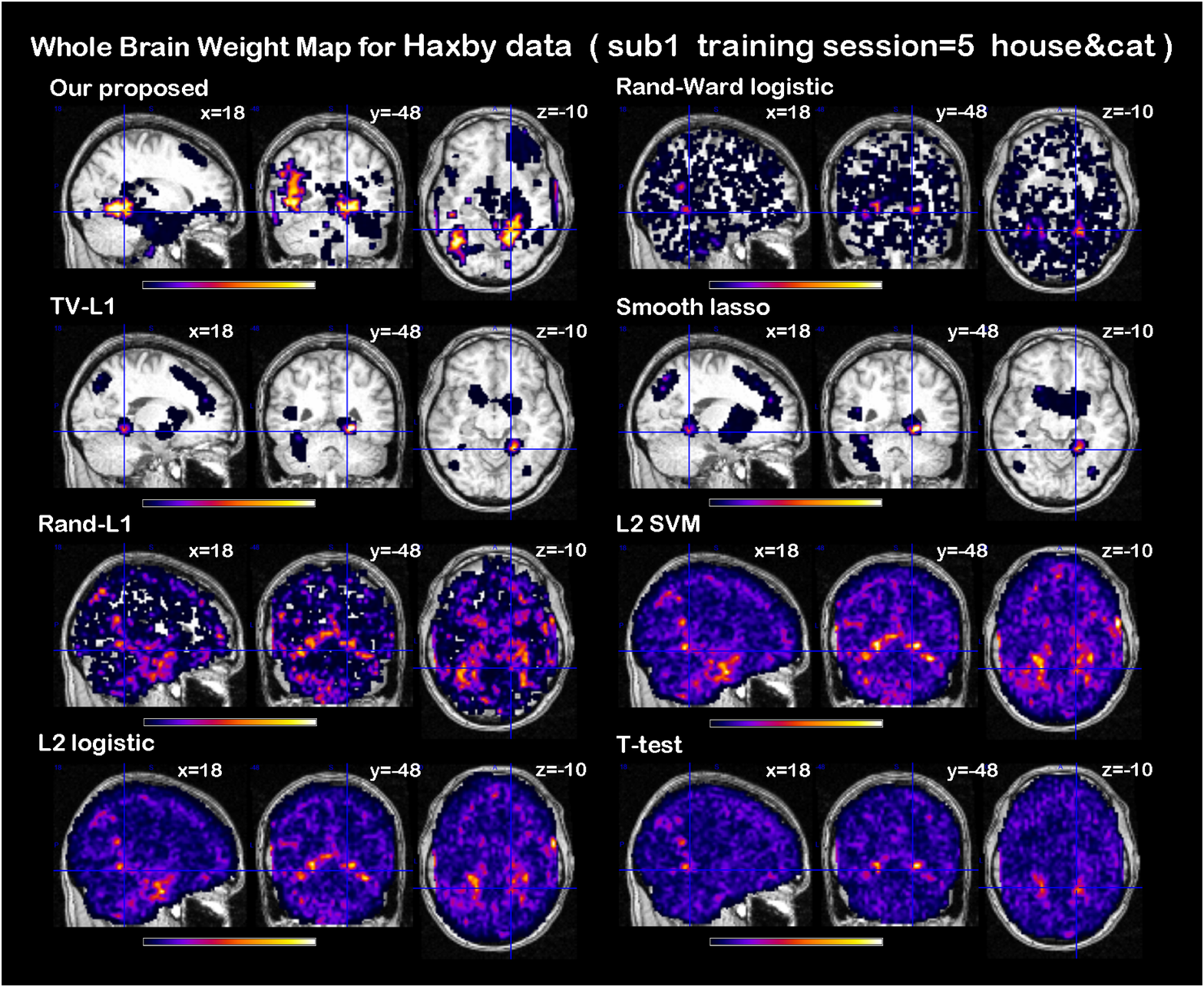}
    \vspace{-0.1cm}\caption{Score maps (unthresholded) as estimated by different methods on fMRI datasets (Cat vs House) using the first $5$ sessions of training data.  Despite being fairly
noisy,  located discriminative brain regions by different algorithm are well highlighted.
     }\label{Fig:haxby_session5_unthreshold}
\end{figure}

\begin{figure}[!h]
    \centering
    \includegraphics[width=0.95\textwidth,height=0.85\textwidth]{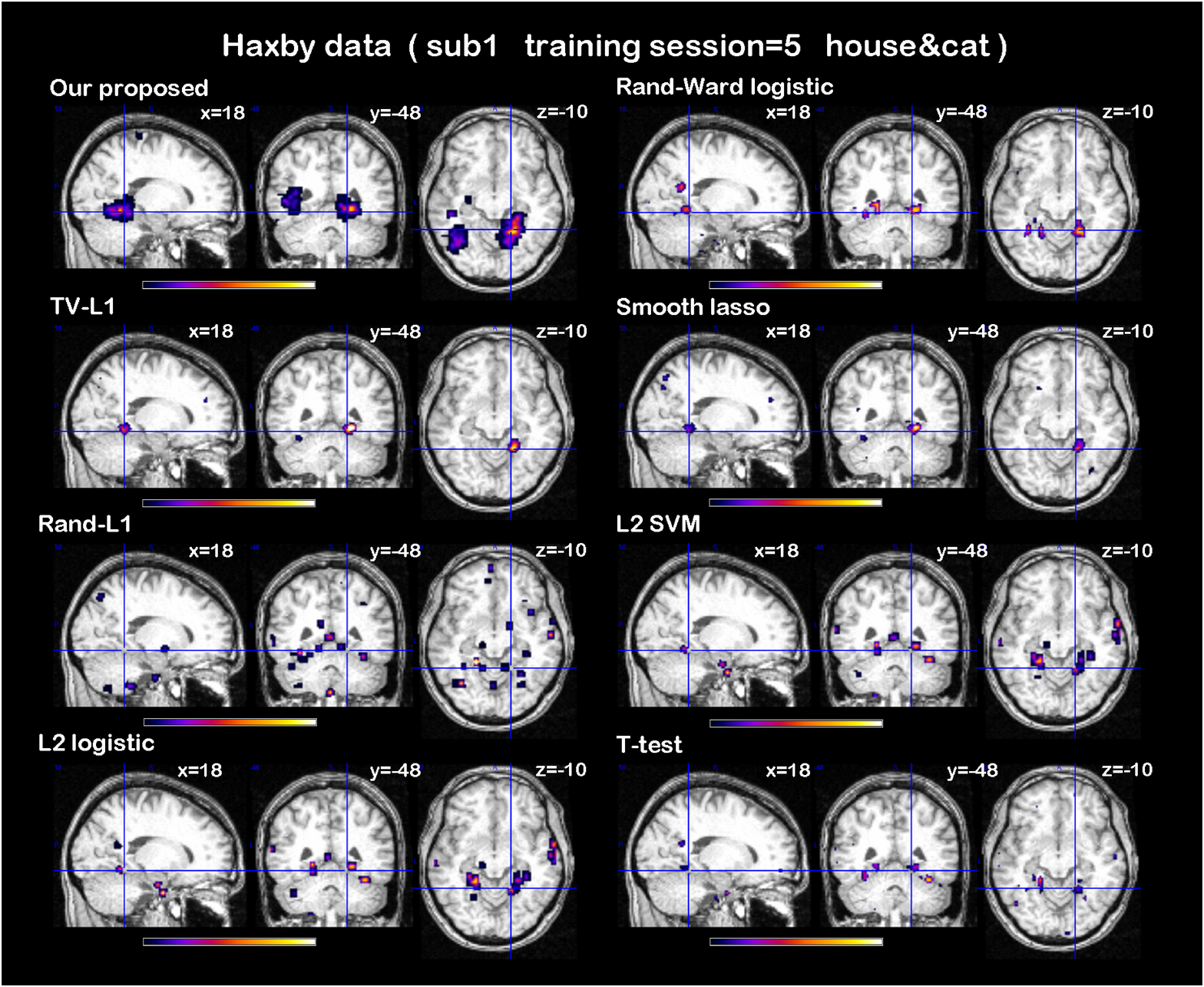}
    \vspace{-0.1cm}\caption{Score maps  as estimated by different methods on fMRI datasets (Cat vs House) using first $5$ sessions of training data.
   The threshold is determined based on cross-validation for the highest prediction accuracy.  Our algorithm can achieve the best performance by finding a larger number of true discriminative voxels than alternatives and keeping the false positives into a very low level. Most of the alternatives have a large number of false positives, except the Randomized Ward Logistic  method, which however, only finds a very small number of true discriminative voxels, although its estimated false positives is $0$.
   }\label{Fig:haxby_session5}
\end{figure}

\begin{figure}[!h]
    \centering
    \includegraphics[width=0.95\textwidth,height=0.85\textwidth]{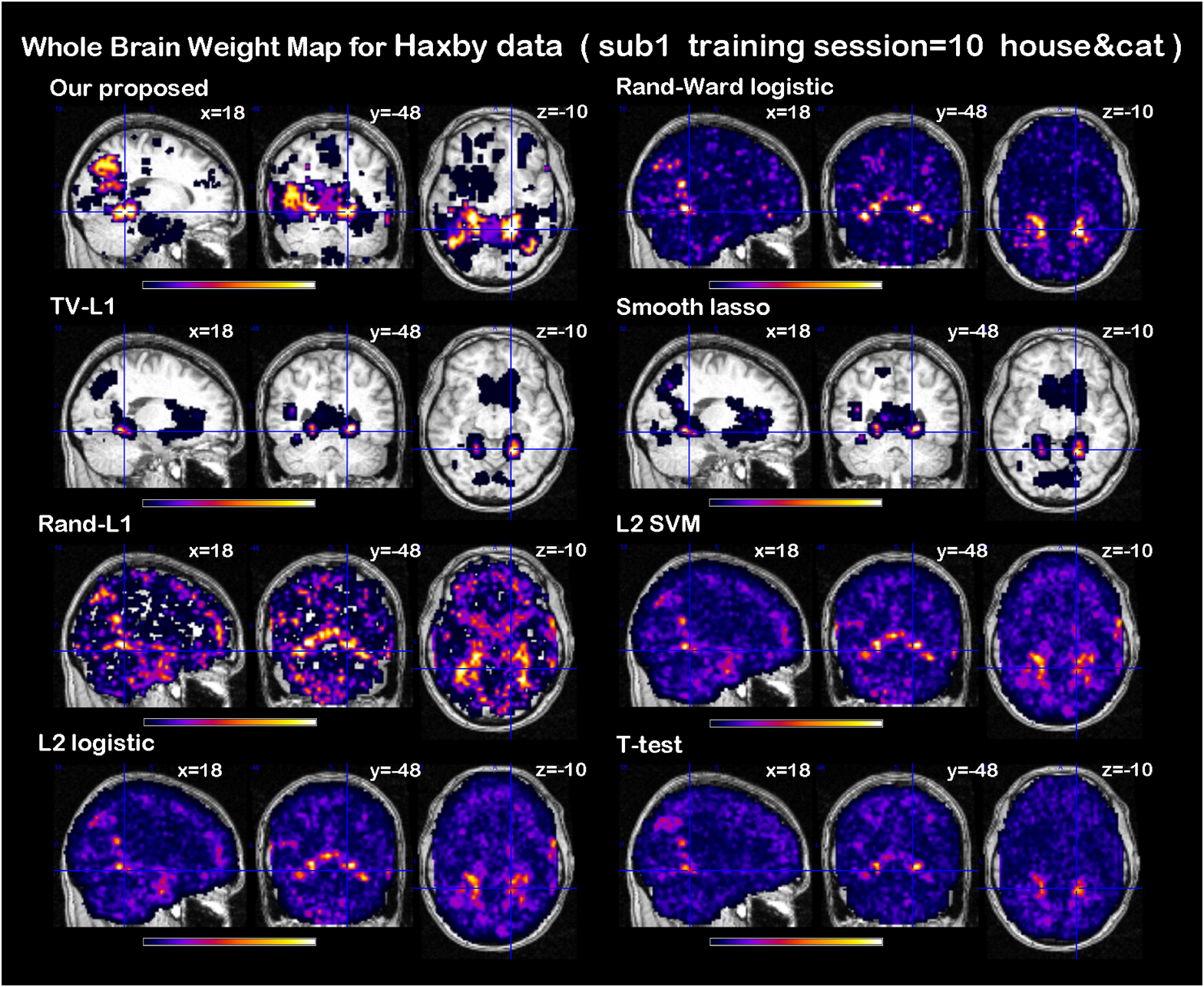}
    \vspace{-0.1cm}\caption{Score maps (unthresholded) as estimated by different methods on fMRI datasets (Cat vs House) using the first $10$ sessions of training data. Despite being fairly
noisy,  located discriminative brain regions by different algorithm are well visually recognized.
   }\label{Fig:haxby_session10_unthreshold}
\end{figure}

\begin{figure}[!h]
    \centering
    \includegraphics[width=0.95\textwidth,height=0.85\textwidth]{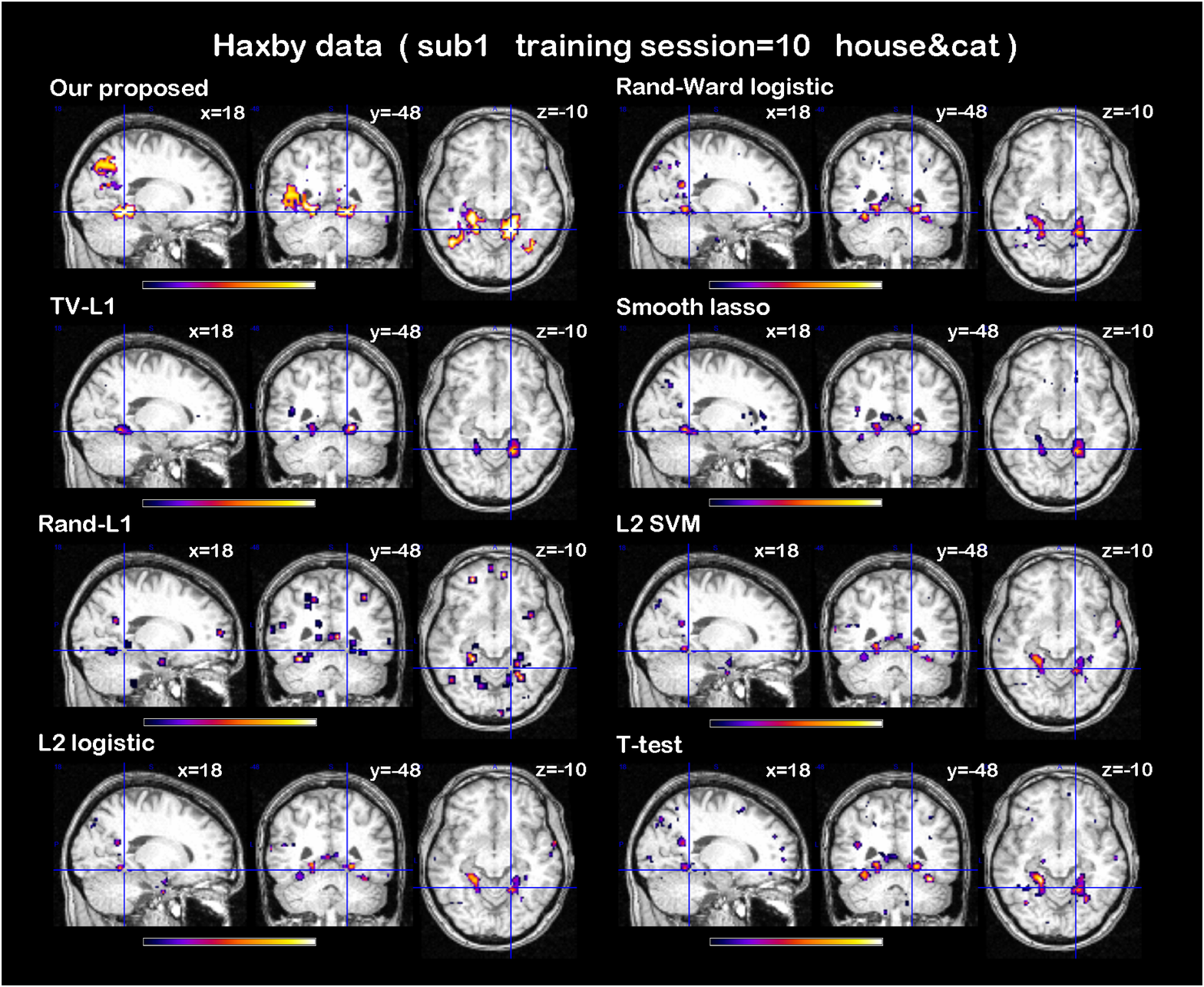}
    \vspace{-0.1cm}\caption{Score maps  as estimated by different methods on fMRI datasets (Cat vs House) using first $10$ sessions of training data.
   The threshold is determined based on cross-validation for the highest prediction accuracy.  Our algorithm can achieve the best performance by finding a large number of true discriminative voxels than alternatives and keeping the false positives into a very low level. Most of the alternatives have a larger number of false positives, except the Randomized Ward Logistic method, which however,  only finds a very small number of true discriminative voxels, although its estimated false positives is $0$.
   }\label{Fig:haxby_session10}
\end{figure}

\begin{figure}[!h]
    \centering
    \includegraphics[width=0.95\textwidth,height=0.65\textwidth]{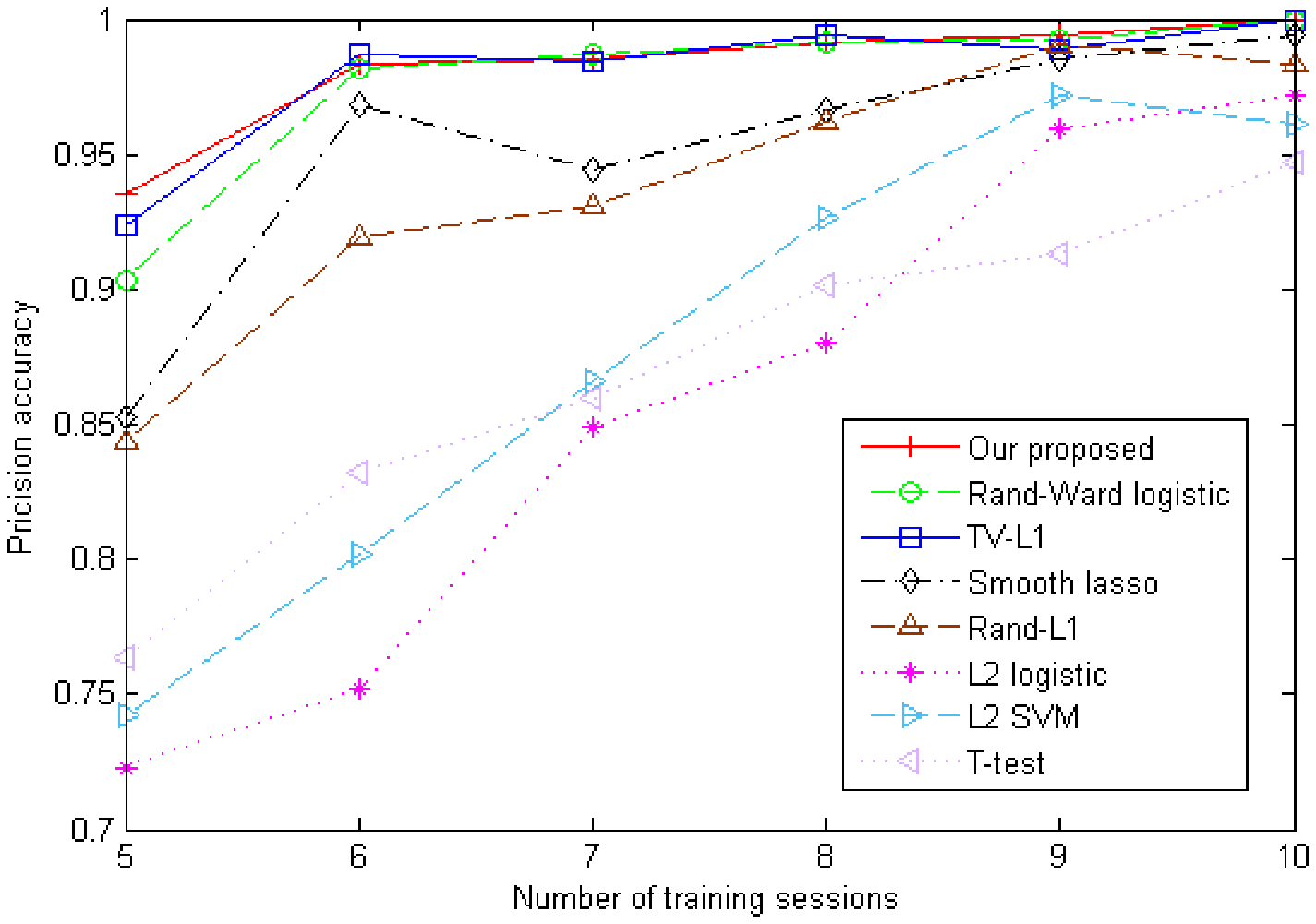}
    \vspace{-0.1cm}\caption{ The classification accuracies of different algorithms when classifying the Cat and House.
Our algorithm is among those achieving the highest prediction accuracy in general. 
   }\label{Fig:haxby_prediction_accuracy}
\end{figure}

\subsection{Real fMRI Data- Haxby Cognitive Task  Data}
We also test our algorithm on a public, block-design fMRI dataset from a study on face and object
representation in human ventral temporal cortex \citep{Haxby01MVA}. The set, which can be downloaded at \url{http://data.pymvpa.org/datasets/haxby2001/}.
consists of $6$ subjects
with $12$ runs per subject. In each run, the subjects passively viewed greyscale
images of eight object categories, grouped in 24s blocks separated by rest
periods. Each image was shown for 500ms and was followed by a 1500ms
inter-stimulus interval.  Full-brain fMRI data were recorded with a volume
repetition time of 2.5s. Then a stimulus block was covered by roughly $9$
volumes.  For a complete
description of the experimental design, fMRI acquisition parameters, and
previously obtained results see the reference \citep{Haxby01MVA}. There is no smoothing operation on this data. In this paper, we consider the fMRI data of the first subject when classifying the ``house" and ``cat", which  consists of $12$ sessions in total. The number of samples, for the first $5$ sessions,  is $90$. The number of training samples evenly increases to $180$ when the number of sessions are $10$.  We adopt the spatially constrained spectral clustering algorithm, implemented in a python software ``PyClusterROI" \citep{Craddock13connectomes}.
 The number of clustering is $200$ when the number of sessions is $5$ and evenly increases to $400$ when the number of sessions is $10$.

We  use the first $T$ sessions as  training samples to perform the feature selection, where $T=5, 6, 7 ,8 ,9, 10$. Then we obtain a prediction score of these selected features,  on $12-T$ remaining  sessions, which are used as the test samples. Due to the limited length of this paper, we only show the brain maps obtained when we use the first $5$ and $10$ sessions as the training data.


Figures \ref{Fig:haxby_session5_unthreshold} and \ref{Fig:haxby_session10_unthreshold} are  brain maps  based on the scores of different methods:  the first $5$ and $10$ sessions, respectively. The scores are not thresholded for visualization purposes. Figures \ref{Fig:haxby_session5} and \ref{Fig:haxby_session10} show the thresholded maps of different algorithms, when we use the first $5$ and $10$ sessions as the training data, respectively. The threshold values of different algorithms can be different. The T-test is based on the common requirement that p-value is less than $0.001$ for FDR control. All the rest of the methods are based on cross-validation, where a linear $\ell_2$-SVM classifier is used and the remaining $7$ or $2$ sessions are used as test data. The candidate threshold value corresponding to its own  best prediction accuracy for each algorithm  is chosen.  Specifically, for our algorithm, we use almost the same threshold-setting procedure as the Python code of the Randomized Ward Logistic algorithm by \citep{Varoquaux12Clustering}.
Basically, we first rank the voxels according to the selection scores and set the thresholds from $0.3$ to $0.9$ with step size of $0.1$ and select the final threshold value corresponding to the high classification accuracy on the testing set (the remaining sessions except those for training). 

We can see that in general, our algorithm is among the most sensitive algorithms. In order to test whether the extra voxels  selected only by our algorithm are of convincing prediction power (necessarily for potentially the true positives), we build an $\ell_2$ logistic regression classification model based on  these extra voxels  (i.e., the discriminative voxels obtained by our proposed method subtracts those selected by other methods, respectively). The prediction accuracy of the resulted classifier are listed below in Table \ref{Table:extra}.
We can see that the extra voxels selected by our method give high classification accuracy, showing that at least part of these extra selected voxels could be the relevant voxels of the task.   In the following paragraph, we try to explain and validate the trueness of the discovered discriminative voxels from the viewpoint of neuroscience.

 \begin{table}
\begin{tabular}{|c|c|c|c|c|}
  \hline
 Algorithm   & Rand Ward Logistic& TV-L1 & Smooth Lasso& Rand L1\\\hline
 Pred. Accu.&  97.22\%& 100\% & 100\%  & 97.22\%\\\hline
Algorithm    & L2-SVM & L2-Logistic& T-test &\\\hline
Pred. Accu. &   94.44\% & 94.44\% &  94.44\%&\\
\hline
\end{tabular}\caption{ The prediction (classification) accuracy of the classifier on the voxels selected by RSS subtract those selected by other methods, respectively. Here the first  $10$ sessions are training data, and the rest $2$ sessions are test data.
}\label{Table:extra}
\end{table}


 Our results of threshold and unthreshold maps  both show the same phenomenon that was described in the original case study  \citep{Haxby01MVA}, where the PPA and FFA are included. The area of mean response regions across all categories, selected by \citep{Haxby01MVA}, is in consistency with the common regions within our different cluster settings in the axial view. In the unthresholded mapping, the contours are quite similar with different numbers of clusters in the leftmost column;  in the thresholded mapping, selected features are near  the same positions.

As a variant of stability selection, our algorithm maintains the finite sample control of false positives. Our algorithm' advantage  is its improved sensitivity of feature selection comparing to other alternatives. Since there is no ground truth for  evaluation, we have tested whether our detected voxels or regions by various  algorithms are stable and unlikely to be false positives. We did this  by adopting the false positive estimate scheme used in \citep{Rondina14}, which is based on a permutation test and cross validation. In Figure  \ref{Fig:haxby_session10}, the result of our algorithm shows a selection of $1247$ voxels with only  $23$ likely to be false positives. While TV-L1 and Smooth Lasso found larger number of discriminative voxels, $2611$ and $1804$, respectively, and  they also have  $881$ and $977$ voxels, respectively, that are likely to be false positives. Even  the original stability selection, i.e., Rand L1,  has $224$ estimated false positives among its selected $1333$ discriminative voxels.  L2-SVM and L2-Logistic  also have around $200$ estimated false positives. T-test has over $150$ false positives. While our algorithm shares some common components with the Randomized Ward Logistic algorithm, the results are quite different. Randomized Ward Logistic method is more conservative in terms of controlling false positives, at least in its default settings. Its selected voxels have no false positives by the false positive estimate scheme. However, it only reveals $116$ discriminative voxels. Its conservation in this case can even  be observed from the unthresholded Figures \ref{Fig:haxby_session10_unthreshold}.  In contrast, our algorithm finds a large number of true discriminative voxels and keeps the false positives into a very low level. The summary of this result is in Table \ref{Table:False}.

\begin{table}
\begin{tabular}{|c|c|c|c|c|}
  \hline
 Algorithm  & Ours & Rand Ward Logistic& TV-L1 & Smooth Lasso\\\hline
 No.Selected&  1247& 116 & 2611  & 1804\\
No.False Positives  & 23 & 0 & 881 & 977\\\hline
Algorithm&    Rand L1 & L2-SVM & L2-Logistic& T-test \\\hline
No.Selected&    1333 & 1499 & 1549& 1045\\
No.False Positives    & 224 & 204 & 198 & 151\\
    \hline
\end{tabular}\caption{ ``No. Selected" means the number of selected voxels after thresholding when using the first $10$ sessions. ``No.False Positives" is the number of probable  false positives among all the selected voxels, estimated via permutation test and cross-validation, as suggested in \citep{Rondina14}.
}\label{Table:False}
\end{table}


Now we look at the predictive power of the best selected voxels of different algorithms. These prediction results are reported in Figure \ref{Fig:haxby_prediction_accuracy}.  We consider to pick  T sessions as the training data and the remaining 12-T sessions are  the test data, where T= $5, 6, 7, 8, 9, 10$. Here we randomly pick T sessions from the 12 sessions and consider all the possible combinations.  The average prediction accuracy among all the combinations for each T  is presented.   Our algorithm is among those that achieve the highest predictive accuracy. Notice that while Randomized Ward Logistic reveals only a small number of discriminative voxels, its prediction accuracy is also very high. While  high predictive accuracy does not directly prove the sensitivity or specificity of  feature selection results, it still suggests the quality of the identification of  voxels of different algorithm to some degree. At the least, the prediction scores suggest that our algorithm does indeed find the relevant voxels because they can achieve significantly high predictive accuracy.

\subsection{A Brief Computational Efficiency Description}\label{runningtime}
All the above experiments were  performed under Windows 7 and MATLAB\_R2014a(V8.3.0.532)
 running on a desktop with  Intel Core i7 Quad-Core (Eight-Thread) Processor with Processor Base Frequency 3.5GHz and $64$ GB of memory, though there are no parallel implementations of all the involved algorithms.  We listed the running time (unit of time is minute here) of different algorithms for the test problems based on both Chess-Master data and Haxby Cognitive Task  Data in Table \ref{Table:time}. Here we did not list the running of Smooth lasso and $\ell_2$-SVM,  $\ell_2$ logistic regression and T-test, because they in general take much shorter time than the listed $4$ algorithms.

Notice that the random ward clustering algorithm is written in Python, while our algorithm is written in MATLAB. Python is in general a more computationally efficient computer language than MATLAB.  So the advantage of our algorithm in terms of computational efficiency comparing with the random ward clustering algorithm is remarkable. Notice that for random ward clustering algorithm, its running time is significant longer for Chess-Master test problem than the Haxby test problem. The number of features of Chess-Master test problem is $91\times 109\times 91$ while the number of features of Haxby  test problem is $40\times 64\times 64$. The spatially constrained ward clustering method used in the random ward clustering algorithm  empirically takes a notably  much longer time as the number of features increases.  While both ours and  random ward clustering algorithm take a longer time than the other alternative algorithms as expected, the running time is still acceptable in general. Finally, we would like to point out that for random ward clustering algorithm and TV-L1 algorithms, we directly use the default settings of the provided python software. Their computational efficiency could be much different if  different parameters are used. So the presented running time of different algorithms here is only for a rough reference. 



\begin{table}
\begin{tabular}{|c|c|c|c|c|}
  \hline
 Algorithm $\rightharpoonup$ & Ours & Rand Ward Logistic& TV-L1 & Rand L1\\\hline
 Haxby&  35 & 68  & 4  & 10\\\hline
Chess-Master  & 52 & 433 & 15 &36\\\hline
\end{tabular}\caption{ Running time (unit: minute) of different methods for second problem (Chess-Master Data) and the third problem  (Haxby Cognitive Task Data, training sessions=10). 
}\label{Table:time}
\end{table}
\section{Conclusion and Future Work} \label{Sec:conclusion}


%

 Voxel selection is very important for decoding fMRI data. 
In this paper we propose a simple and computationally efficient method for data-driven
voxel selection which is also called support identification,  for  potential biomarker extraction \citep{Orru2012SVM}. We propose a ``\textit{randomized structural sparsity}" as a structural variant of  classical stability selection via  specific implementation ``\textit{Constrained Block Subsamplings}". We apply this to the existing sparse multi-variate classifiers such as $\ell_1$ logistic regression, in the case of fMRI data, which has  strong correlations and distributed multivariate discriminative patterns.
However, the results are mostly empirical and we might need to perform theoretical support in order to better understand its advantages and address its limitations. For example, the theoretical results about the false positive rate and false negative rate of our feature selection  algorithm need to be presented in the future work. 
%
%
In addition,  we need to further study the possible bias or arbitraries brought by our one-time parcellation. Moreover, we would like to try the hierarchical ward clustering with spatially constraints in our algorithm in future. It  has showed that in general Ward's clustering performs better than alternative methods with regard to reproducibility and accuracy \citep{Thirion14Clustering}. Furthermore, how to effectively distinguish true positives from false positives needs to be better addressed.

\section*{Acknowledgment}
Thanks to Prof. Alexandre Gramfort of Telecom ParisTech  for kindly providing us with the Python Smooth LASSO and TV-L1 code, which is  under integration in the Nilearn package. Thanks to Dr. Ga\"{e}l Varoquaux from  Parietal team, INRIA, for kindly providing us with the Randomized Ward Logistic algorithm written in Python. We would also like to thank the anonymous reviewers for their many constructive  suggestions, which have greatly improved this paper.


This work was supported by the 973 programs (Nos. 2015CB856000, 2012CB517901), 863 project (SQ2015AA0201497), the Natural Science Foundation of China (Nos. 11201054, 91330201,61125304, 81301279), and the Specialized Research Fund for the Doctoral Program of Higher Education of China (No. 20120185110028)£¬and, the Fundamental Research Funds for the Central Universities(ZYGX2013Z004,ZYGX2013Z005)

\bibliographystyle{model2-names}
\bibliography{GroupSparsity,Pattern_recognition,brainScience,StabilitySelection}







\end{document}